%% file: main.tex
\documentclass[conference]{IEEEtran}
\usepackage[nolist,nohyperlinks]{acronym}

\ifCLASSINFOpdf
  \usepackage[pdftex]{graphicx}
\else
  \usepackage[dvips]{graphicx}
\fi
\usepackage{units}
\usepackage{amsmath}
\usepackage{textcomp}          

\begin{document}
%
\title{Wireless Interference Identification with\\ Convolutional Neural Networks}

\author{\IEEEauthorblockN{Malte Schmidt, Dimitri Block, Uwe Meier}
\IEEEauthorblockA{inIT - Institute Industrial IT\\
	Ostwestfalen-Lippe University of Applied Sciences\\
	Lemgo, Germany\\
Email: malte.schmidt@rwth-aachen.de, dimitri.block@hs-owl.de, uwe.meier@hs-owl.de}}


%


\maketitle

\begin{abstract}
The steadily growing use of license-free frequency bands requires reliable 
coexistence management for deterministic medium utilization. 
For interference mitigation, proper \ac{WII} is essential.

In this work we propose the first \ac{WII} approach based upon deep \acp{CNN}.
The \ac{CNN} naively learns its features through self-optimization 
during an extensive data-driven GPU-based training process.
We propose a \ac{CNN} example which is based upon sensing snapshots 
with a limited duration of \unit[12.8]{\textmu s} 
and an acquisition bandwidth of \unit[10]{MHz}.
The \ac{CNN} differs between 15 classes.
They represent packet transmissions of 
 IEEE 802.11 b/g, IEEE 802.15.4 and IEEE 802.15.1
with overlapping frequency channels within the \unit[2.4]{GHz}~ISM~band.
We show that the \ac{CNN} outperforms state-of-the-art \ac{WII} approaches 
and has a classification accuracy greater than \unit[95]{\%} for \acl{SNR} 
of at least \unit[-5]{dB}.
\end{abstract}


%
\IEEEpeerreviewmaketitle

\acresetall
\input{conf_introduction}
\input{conf_related_work}
\input{conf_data_set}
\input{conf_cnn}
\input{conf_results}

\acresetall
\input{conf_conclusion}

\bibliographystyle{IEEEtran}
\bibliography{literatur}

\begin{acronym}[Bash]
	\acro{ReLU}{rectified linear unit}
	\acroplural{ReLU}[ReLUs]{\emph{rectified linear units}}
	
	\acro{SGD}{stochastic gradient descent}
	
	\acro{PSD}{power density spectrum}
	
	\acro{CNN}{convolutional neural network}
	\acroplural{CNN}[CNNs]{convolutional neural networks}
	
	\acro{SNR}{signal-to-noise ratio}
	
	\acro{GPU}{graphics processing unit}
	
	\acro{NFSC}{neuro-fuzzy signal classifier}
	
	\acro{VSG}{vector signal generator}
	
	\acro{RSA}{real time spectrum analyzer}
	
	\acro{CPU}{central processing unit}
	
	\acro{MNIST}{Mixed National Institute of Standards and Technology}
	
	\acro{WII}{wireless interference identification}
	
	\acro{EER}{equal error rate}
	
	\acro{FFT}{fast fourier transform}
\end{acronym}

\end{document}

%% file: conf_introduction.tex
\section{Introduction}

Artificial neural networks and especially \acp{CNN} achieved excellent results
for different benchmarks in recent 
years~\cite{2012.CiresanMeierSchmidhuber,2012.KrizhevskySutskeverHinton,2013.LinChenYan}.
Neural networks achieve the best performance e.g. for character recognition of the
\ac{MNIST} database~\cite{2016.LecunCortes}. The results achieved by the 
\acp{CNN} from Cire\c{s}an et al.~\cite{2012.CiresanMeierSchmidhuber} are 
comparable to human performance. Inspired by these results and the progress
in deep learning, \acp{CNN} are used as classifier in a growing number of
research fields.

One of these research fields is \ac{WII} for coexistence management of license-free 
frequency bands such as the \unit[2.4]{GHz}~ISM~band.
Such bands are shared between incompatible heterogeneous wireless communication systems. 
In industrial environments, typically standardized wireless communication 
systems within the \unit[2.4]{GHz}~ISM~band are wide-band high-rate IEEE 802.11 
b/g/n, narrow-band low-rate IEEE 802.15.4-based WirelessHART and ISA 100.11a, 
and IEEE 802.15.1-related PNO WSAN-FA and Bluetooth. 
Additionally, the spectrum band is shared with many proprietary wireless technologies which target specific application requirements such as the IEEE 802.11-based industrial WLAN (iWLAN) from Siemens AG, FHSS-based Trusted Wireless from Phoenix Contact and IEEE 802.15.1-based WISA from ABB Group.

Any radio interference can cause packet loss and transmission latency for 
industrial radio communication systems.
Both effects have to be mitigated for real-time medium requirements.
Therefore, the norm IEC 62657-2 \cite{2013.IEC626572:2013} for industrial radio communication systems recommends an active coexistence management for reliable medium utilization.
\cite{2013.IEC626572:2013} recommends (i) manual, (ii) automatic 
non-cooperative or (iii) automatic cooperative coexistence management.
The first approach is the most in-efficient one, due to time-consuming complex configuration effort.
The automatic approaches (ii) and (iii) enable efficient self-reconfiguration without manual intervention and radio-specific expertise.
An automatic cooperative coexistence management (iii) requires a control channel, i.e. a logical common communication connection between each coexisting wireless system to enable deterministic medium access.
In case of a single legacy coexisting wireless system without such connection, the non-cooperative approach (ii) is recommended.
Non-cooperative coexistence management approaches are aware of coexisting wireless systems based on independent \ac{WII} and mitigation.

In this paper, we propose the first \ac{WII} approach based upon deep \acp{CNN}.
In order to face realistic wireless device capabilities the approach 
is limited to a sensing bandwidth of \unit[10]{MHz}
and a sensing snapshot is limited to \unit[128]{IQ Samples} with the duration 
of \unit[12.8]{\textmu s}.
The evaluation is performed with standardized wireless communication systems
based upon IEEE 802.11 b/g, IEEE 802.15.4 and IEEE 802.15.1, 
which are sharing the \unit[2.4]{GHz}~ISM~band. 
In total 19 different variants of modulation types and symbol rates are utilized.
Thereby, the \ac{WII} approach has to differ between 15 classes 
which represent the allocated frequency channel and the wireless technology.

First of all in chapter~\ref{sec:related_work} we will present and discuss 
three 
publications related to our
work: Compressed Sensing, the \acl*{NFSC} and Convolutional Modulation
Recognition. Then, in 
chapter~\ref{sec:data_set}, we will explain how we generated our data set and 
discuss our \ac{CNN} design in chapter~\ref{sec:cnn_design}. Thirdly in 
chapter~\ref{sec:results} we will evaluate the performance of our \acp{CNN} and 
compare it to the performance of the \ac{NFSC}. Finally in 
chapter~\ref{sec:conclusion} we will suggest future work.

%% file: conf_related_work.tex
\section{Related Work}
\label{sec:related_work}

\subsection{Compressed Sensing}
Compressed sensing is utilized for sub-sampled signal reconstruction.
In \cite{2007.TianGiannakis, wieruch2016cognitive} it is also applied for classification of  
frequency bands into white, gray and black sections. Such separation depends on 
the spectral power density and shows in which sections of the frequency 
band a transmission without interference is likely.

This method tries to reduce the run-time of the decision process as much as 
possible. Therefore, the computation of the \ac{PSD} is done with a compressed 
vector of input samples for the trade-off of accuracy. The goal of this method
is not to exactly classify which radio signals are present but to find
spectral spaces which are sparely used. For this task a wavelet-based edge
detector is used.

In contrast to our approach the compressed sensing promises faster 
run-times and can make use of samples recorded with a sample rate lower than
the Nyquist rate. This means lower hardware requirements. On the other hand this
leads to an information loss which might be helpful for decision
making.

\subsection{Neuro-Fuzzy Signal Classifier}
The \ac{NFSC}~\cite{2010.Ahmad} is an expert system which classifies frequency 
bands with respect to known wireless technologies. The purpose of the \ac{NFSC} 
is similar to the proposed \ac{CNN}. The performance of the two classifiers 
will be compared later on.

The \ac{NFSC} is separated into six different layers. The input of the first 
layer, the input layer, are the IQ samples of the signal. This layer computes 
the \ac{PSD} of the input. 

In the second layer, the fuzzification layer, the 
\ac{PSD} is normalized to
\begin{equation}
\label{equ:normalization}
\mu[n] = \left| \frac{\operatorname{min}\left( P\right)  - P\left[ n\right]}
{\operatorname{min}\left( P\right)  - \operatorname{max}\left( P\right) } 
\right| 
\end{equation}
with the \ac{PSD} $P$ in \unit{dBm}.

In the third layer, the filtering layer, the signal is filtered with a 
predefined filter shape. In the simplest case the filter shape is a rectangle 
that is greater than the channel bandwidth of the radio signal that has to be 
classified.

After the filtering layer the similarity layer computes the similarity of the 
filtered signal with a predefined reference shape. After comparing the 
similarity to a threshold the radio signal is classified. Given the reference 
shape $R$ and the fuzzificated, filtered signal $S$ the similarity $SM$ is 
given by
\begin{equation}
\label{equ:similarity}
\textit{SM} = \frac{\sum\limits_n  \operatorname{min}\left( R\left[ n\right] , 
S\left[ n\right]  \right) }
{\operatorname{max}\left(\sum\limits_n R \left[ n\right] ,  \sum\limits_n S 
\left[ n\right] \right)} \text{.}
\end{equation}

Thus it is possible to define many different filter and reference shapes to
classify different radio signals at the same time. 

The fifth layer, the statistics layer aggregates consecutive results for 
temporal evaluation.
This information is handed over to the sixth layer, the interference layer,
which decides if the radio signal belongs to a frequency-hopping system. 

In this paper the \ac{NFSC} up to the fourth layer is compared to the \ac{CNN} 
because the \ac{CNN} has only a very short measurement period and temporal 
statistics must be 
collected by a following processing unit. 

The purpose of the \ac{NFSC} is exactly the same as the \ac{CNN} presented in 
this paper: to classify radio signals with respect to known standards. The 
approach 
however is quite different. While the \ac{NFSC} relies on pre-defined features 
and a fixed decision process the \ac{CNN} trains its feature extraction and 
decision process during a learning process. On the one hand this leads to a
more flexible decision making. On the other hand it becomes more difficult to
analyze the decision making process.

\subsection{Convolutional Modulation Recognition}
O'Shea et al.~\cite{2016.OSheaCorgan} compared the performance of different 
classifiers for classifying 11 modulations, 8 digital and 3 analog modulations. 
The classifiers used were \acp{CNN} with different parameters, deep neural 
networks with different parameters, a decision tree, a naive bayes classifier,
a k-nearest-neighbor classifier and a support vector machine. For this task a 
\ac{CNN} achieved the best results for low \acp{SNR}. For high \acp{SNR} the 
difference of all classifiers were comparable except the naive bayes and a deep
neural network with high regularization.

O'Shea et al. investigated the complex-valued temporal radio signal domain 
whereas in this paper the spectral radio signal domain is investigated. 
Nevertheless there are many parallels in these domains and we 
adapted the \ac{CNN} from O'Shea et al. that achieved the best results as a 
starting point for our work. This also shows how flexible such self-learning 
classifiers can be used. The network structure of this \ac{CNN} was inspired by 
networks for the visual domain like the \ac{MNIST} data set.

For data set generation O'Shea et al. used GNU 
Radio~\cite{2001.Blossom} a 
toolkit for software 
defined radios. The \ac{CNN}  was trained on approximately 96,000 snapshots 
each consisting of \unit[128]{IQ samples}. These IQ samples were given as a 
$128 
\times 2$ matrix as the input for the \ac{CNN}. One column of the matrix
consisted of the I-samples and the other column of the Q-samples. No
information about the link between the I- and Q-samples were given to the
network. The snapshots were snapshots of the time domain. The data 
set is available at \cite{2016.radioml}.

%% file: conf_data_set.tex
\section{Generation of Data Set}
\label{sec:data_set}
The training and validation data that was used was generated with the \ac{VSG} 
SMBV100A from RHODE \& SCHWARZ and was recorded with the \ac{RSA} RSA6114A from
TEKTRONIX. 

The measurement was triggered. Only data was recorded in which a radio
signal was transmitted. For example the inter-frame gaps of the signals were
not recorded. Every package that was sent by the \ac{VSG}
had the maximum allowed number of bytes as payload. The payload was chosen at
random. 

For the IEEE~802.11 b/g frames the Physical Layer Mode was varied between CCK,
PBCC and OFDM and every allowed bit rate for each mode was used. For the
IEEE~802.15.1 frames the Transport Mode was varied between ACL, eSCO and SCO
and different packet types were used. For the IEEE-802.15.4 frames the
ACK-frame was used.

In the presented work some restrictions concerning the training and validation
data were made:
\begin{itemize}
	\item Single-label classification
	\item Flat fading channel model
	\item Thermal noise reception distortions
\end{itemize}

A single-label problem is considered, so signals of exactly one
class are present in each input sensing snapshot for the \ac{CNN}. This means 
no concurrent signals were allowed. 
Moreover for the generating of the data only one emitter, the \ac{VSG}, was used. 

Further, a flat fading channel was utilized due to the
connection of the \ac{VSG} and \ac{RSA} via a coaxial cable. These restrictions
were made to keep the problem simple to get a first prototype running.

The third restriction is an assumption of thermal noise reception distortions
with additive white Gaussian noise 
in the \ac{SNR} range of \unit[-20]{dB} until \unit[20]{dB} with the step size of \unit[2]{dB}.
It was added with a SIMULINK~\cite{2016.Matlab} model in post-processing.

In total 151,200 sensing snapshots were used for training and 74,025 for 
validation.

%% file: conf_cnn.tex
\section{Convolutional Neural Network Design}
\label{sec:cnn_design}
The \acp{CNN} design targets radio signal classification. 
The radio signals are complaint to IEEE 802.11 b/g, IEEE 802.15.4 and IEEE 802.15.1 packet transmissions.
Thereby, the \ac{CNN} shall classify the allocated frequency channel 
and the corresponding wireless technology.

In order to face realistic wireless device capabilities the \ac{CNN} 
is limited to a sensing bandwidth of \unit[10]{MHz}.
Hence, to observe the whole \unit[2.4]{GHz}-ISM-Band eight parallel classifiers are required.
Therefore, a technology-specific relative channel number $RO_\text{WT}$ can be mapped 
to its absolute channel number $ACH_\text{WT}$
with the index $n_\text{CNN} \in \{1, 2, ..., 8\}$ of the utilized \ac{CNN}:
\begin{equation}
ACH_\text{WT} = RCH_\text{WT} + RO_\text{WT} \cdot (n_\text{CNN} -1) + AO_\text{WT}
\end{equation}
with technology-specific absolute and relative channel offsets $RO_\text{WT}$ and $AO_\text{WT}$, respectively.
The offset values and the channel number sets are listed in Tab. \ref{tab:wtch}.
The limited sensing bandwidth comprises ten, two, and three frequency channels 
of IEEE 802.15.1, IEEE 802.15.4, and IEEE 802.11 b/g complaint signals, respectively.
In total, $15$ different classes have to be distinguished.

\begin{figure}[htb]
	\centering
	\includegraphics[width=\columnwidth]{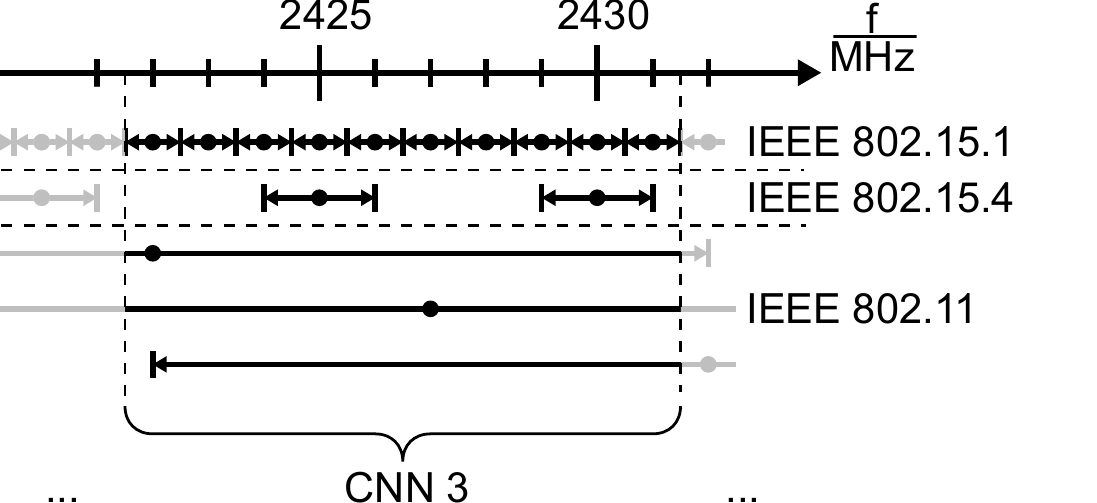}
	\caption{Frequency channel classes of example \ac{CNN} in the 
	\unit[2.4]{GHz}-ISM-Band.}
	\label{fig:konzept10MHz}
\end{figure}

Fig.~\ref{fig:konzept10MHz} illustrates the classes which represent
frequency channels of the corresponding wireless technologies 
in case the third \ac{CNN} with the center frequency of \unit[2426.5]{MHz} is utilized.
It is important to mention, 
that the frequency channels of the selected IEEE 802.15.1 and IEEE 802.15.4 complaint signals are within sensing bandwidth
while the signals complaint to IEEE 802.11 b/g are only partly within sensing bandwidth.

\begin{table*}[htb]
\centering
\caption{Wireless Technologies Channel Mapping}
\label{tab:wtch}
\begin{tabular}{llccc}
\hline
\textbf{Wireless technology} & WT & \textbf{IEEE 802.15.1} & \textbf{IEEE 802.15.4} & \textbf{IEEE 802.11} \\
\hline
\textbf{Absolute channel numbers} & $ACH_{\text{WT}}$ & $1, 2, \ldots, 79$ & $1, 2, \ldots, 15$ & $1, 2, \ldots, 13$ \\
\hline
\textbf{Relative channel numbers} & $RCH_{\text{WT}}$ & $0, 1, \ldots, 9$ & $0, 1$ & $0, 1, 2$ \\
\hline
\textbf{Absolute offset} & $AO_{\text{WT}}$ & $1$ & $1$ & $1$ \\
\hline
\textbf{Relative offset} & $RO_{\text{WT}}$ & $10$ & $2$ & $2$ \\
\hline
\end{tabular}
\end{table*}

The sensing bandwidth is exemplary and other bandwidths are possible.
A greater sensing bandwidth with a fixed sensing snapshot duration increase the input data size proportionally. 
Additionally, the number of observable frequency channels and therefore distinguishable classes increases.
Therefore, the \acp{CNN} requires more neurons in at least the first and last 
layer, 
which increases the computation complexity and also the required number of 
IQ-samples for a sensing snapshot.

\subsection{Frequency-Domain Sensing Snapshots}
The sensing snapshot is limited to a duration of \unit[12.8]{\textmu s} 
and therefore consists of \unit[128]{IQ-Samples}.
It has to be greater than the symbol durations of the utilized wireless technologies 
for sense-making classification.
With the minimal applied symbol duration of \unit[1]{\textmu s}, 
a single sensing snapshot contains up to \unit[12.8]{symbols}.

Danev et al. \cite{2009.DanevCapkun} showed for emission-based wireless device identification
that frequency-based features outperform their time-based equivalents.
Therefore, the IQ-Samples are transfered into frequency-domain by \ac{FFT}.
The resulting snapshot contains 128 complex-valued frequency bins.

\subsection{Network Structure}
The network structure of the \ac{CNN} 
is derived from O'Shea et al.~\cite{2016.OSheaCorgan} as listed in Tab.~\ref{tab:big-cnn}.

Therefore, the input data is a sensing snapshot with complex-valued frequency bins, 
whereby the real and imaginary parts are considered as independent floating point values.
So, the input data results as $128 \times 2$ matrix.

The output data size is a vector of the length 15 as there are 15 classes to
classify. Each entry of the vector has a value between 0 and 1 and describes
how likely it is that the input data belongs to the class it stands for in 
relation to the other classes.

\begin{table}[htb]
\centering
\caption{\ac{CNN} Structure}
\label{tab:big-cnn}
\begin{tabular}{llll}
\hline
\textbf{Layer type}& \textbf{Input size}& \textbf{Parameters} & \textbf{Activation fct.}\\
\hline 
Convolutional & $128 \times 2$ & $3 \times 1$ filter kernel & Rectified linear\\
layer		      &                & $64$ feature maps & \\
\hline
Convolutional & $64 \times 126 \times 2$ & $3 \times 2$ filter kernel & Rectified linear\\
layer        &                          & $1024$ feature maps & \\
        &	                         & Dropout $\unit[60]{\%}$ & \\
\hline
Dense layer 	  & $126976 \times 1$ & $128$ neurons & Rectified linear\\
        &	                  & Dropout $\unit[60]{\%}$ & \\
\hline
Dense layer  	  & $128 \times 1$    & $15$ neurons  & Softmax\\
\hline
\end{tabular}
\end{table}

\subsection{Network Training}

For the training process the Adam optimizer~\cite{2014.KingmaBa} was used and 
the input data was normalized. This optimizer showed best results in the work 
of O'Shea et al.. The default parameters for the Adam optimizer except the
learning rate were used. The learning rate for the \ac{CNN} was $0.0001$. The 
\ac{CNN} was trained for $50$ epochs. This setup showed best
results among small parameter variations. No hyperparameter optimization was
performed but could be done in the future to optimize the results. A batch size 
of $1024$ was used for training which was near the limit of the graphics card 
memory.

\subsection{Network Size Reduction}

As a rule of thumb it is often assumed that the degrees of freedom of the 
\ac{CNN} should be less than the number of sensing snapshots for training. In 
practice it is 
difficult to apply to this rule because it is e.g. not applicable if you also
have to determine the number of sensing snapshots you will use for training. 

As the size of the data set was determined by orientating 
to the work of O'Shea et al.~\cite{2016.OSheaCorgan} we then reduced the size 
of the \ac{CNN} so that it had approximately the same degrees of freedom as 
sensing snapshots were used for training. The idea of this rule is that the 
\ac{CNN} learns to extract and recognize more reliable features in the sensing 
snapshots and does not try to learn random processes like noise in the data. 
This shall lead to a better generalization of the \ac{CNN}. The network 
structure of the reduced \ac{CNN} is listed in Tab.~\ref{tab:small-cnn}.

The learning rate of the reduced \ac{CNN} was $0.001$ and it was trained for 
$200$ epochs. A batch size of $1024$ was used.

\begin{table}[htb]
\centering
\caption{\ac{CNN} Structure with Reduced Network Size}
\label{tab:small-cnn}
\begin{tabular}{llll}
\hline
\textbf{Layer type}& \textbf{Input size}& \textbf{Parameters} & \textbf{Activation fct.}\\
\hline 
Convolutional & $128 \times 2$ & $3 \times 1$ filter kernel & Rectified linear\\
layer &                & $8$ feature maps & \\
\hline
Convolutional & $8 \times 126 \times 2$ & $3 \times 2$ filter kernel & Rectified linear\\
layer &                          & $16$ feature maps & \\
&	                         & Dropout $\unit[60]{\%}$ & \\
\hline
Dense layer   	  & $1984 \times 1$ & $64$ neurons & Rectified linear\\
&	                  & Dropout $\unit[60]{\%}$ & \\
\hline
Dense layer  	  & $64 \times 1$    & $15$ neurons  & Softmax\\
\hline
\end{tabular}
\end{table}

\subsection{Implementation}
The \ac{CNN} was implemented, trained and validated on high end computation platform
with the \ac{CPU} Intel XEON E5-1660 v3, \unit[16]{GBit} RAM and the \ac{GPU} 
Nvidia GTX 960.
The \ac{CNN} implementation utilizes the abstract modular deep learning library 
Keras~\cite{2015.Chollet} and the computation library  
Theano~\cite{2016.Theano} as its backend. 

Performing \ac{CNN}'s training on the \ac{GPU} results in a duration 
of approximately \unit[74]{s} per epoch and therefore \unit[501]{ms} per 
batch. For the reduced \ac{CNN} the duration cuts down 
to approximately \unit[3]{s} per epoch and therefore cuts down to \unit[20]{ms} 
per batch.

%% file: conf_results.tex
\section{Results}
\label{sec:results}

\begin{figure}[htb]
	\centering
	\includegraphics[width=\columnwidth]{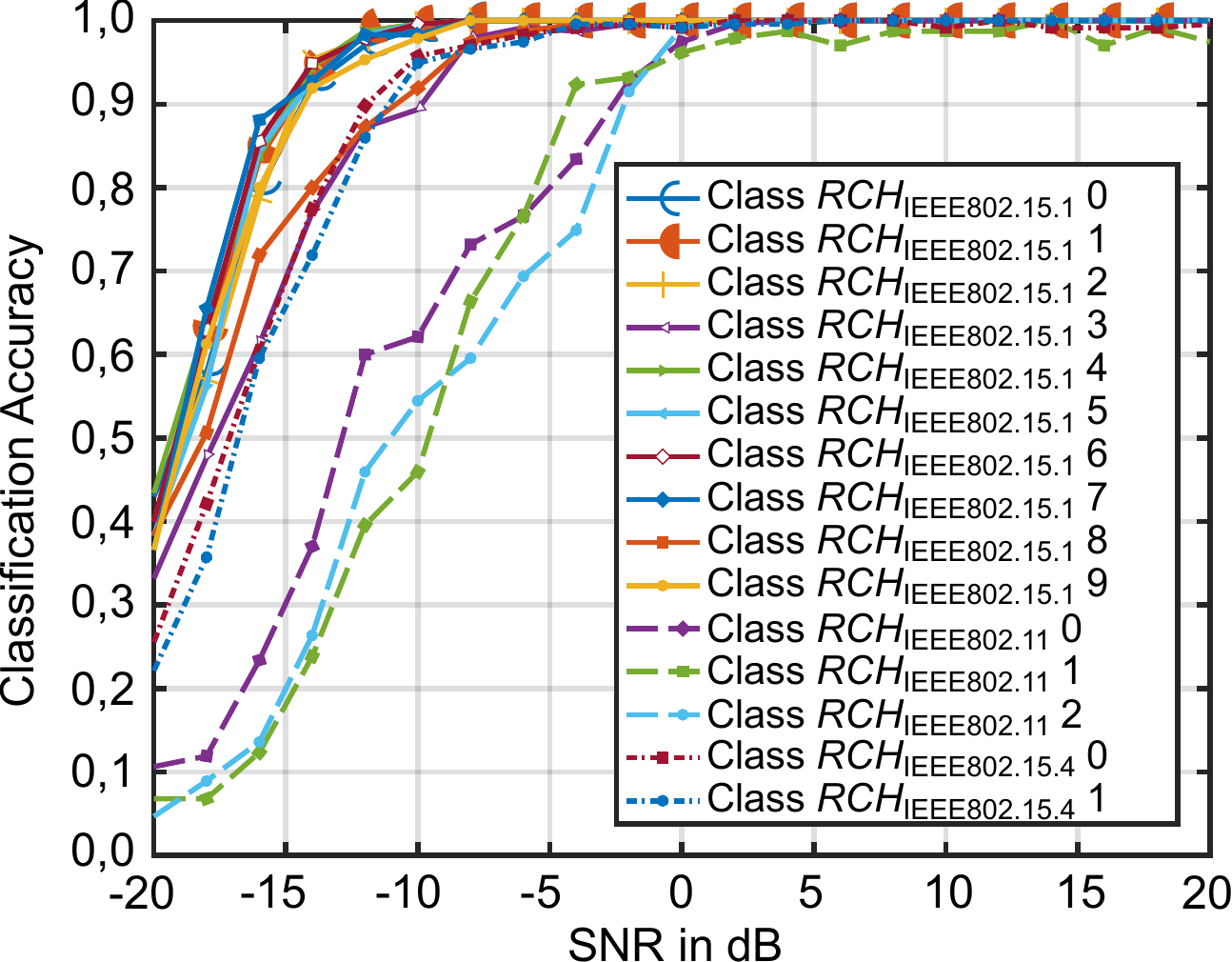}
	\caption{Classification accuracy with varying \acl{SNR}.}
	\label{fig:accuracy_vs_snr_big_cnn_all_classes}
\end{figure}

The accuracy of the \ac{CNN} for the validation data is shown in
Fig.~\ref{fig:accuracy_vs_snr_big_cnn_all_classes}. For the IEEE-802.11 b/g
channels the accuracy is the worst. The IEEE-802.11 b/g signals have the 
biggest
channel width and the most different modulations and bit rates. Therefore they
are the most complex signals for the investigated classification problem.

The best accuracy is achieved for the different IEEE-802.15.1 channels. The
four classes $RCH_\text{IEEE802.15.1}~0, 3, 8$ and $9$ have a slightly worse 
accuracy then the other
IEEE-802.15.1 classes. The channels at the borders of the observed bandwidth,
the classes $RCH_\text{IEEE802.15.1}~0$ and $9$, are deformed by the 
anti-aliasing filter and the channel
bandwidth of the class $RCH_\text{IEEE802.15.1}~9$ overlaps with the channel 
bandwidth of the class $RCH_\text{IEEE802.15.4}~1$.
The classes $RCH_\text{IEEE802.15.1}~3$ and $8$ have the same center frequency 
as the classes $RCH_\text{IEEE802.15.4}~0$ and $1$. This causes a slightly worse
recognition rate by the \ac{CNN} because there are less features to distinguish
between these classes.

The accuracy for all
\acp{SNR} is clearly better than the accuracy achieved by the \ac{CNN} for
modulation recognition used by O'Shea et al.~\cite{2016.OSheaCorgan}. There are
two possible reasons for this result. The first possible reason is that the
training and validation data was generated too synthetically and does not 
represent the real world effects well enough. The other possible reason which
seems to be more likely is that a mainly frequency selective
classification is much easier for a \ac{CNN} to classify. Another hint which
leads to this conclusion is that the \ac{CNN} could be reduced significantly
without a big loss of accuracy. 

The significant reduction of the \ac{CNN} points out that a frequency selective 
classification problem is less complex than a modulation selective 
classification which was investigated by O'Shea et 
al.~\cite{2016.OSheaCorgan}.  Another proof for this assumption is the slightly 
worse recognition rate of the IEEE~802.15.1 signals which have the same
center frequency as the IEEE~802.15.4 signals. If all signals that have to be 
classified use the same center frequency the accuracy will likely get worse as 
it is the case in the work of O'Shea et al..

\subsection{Comparison of \ac{CNN} and \ac{NFSC}}

The accuracy of the \ac{CNN} was compared to the accuracy of the \ac{NFSC}. 
Therefore, the filter and reference shape of the \ac{NFSC} rectangles were used. The
filter shape was twice as wide as the reference shape which had the channel
bandwidth as width. 
Further, a threshold of $0.5$ for the similarity was used. 
For fair comparison, the classification of IEEE-802.11 b/g complaint signals were ignored 
due to suboptimal performance for out-of-band signal identification.

\begin{figure}[htb]
	\centering
	\includegraphics[width=\columnwidth]{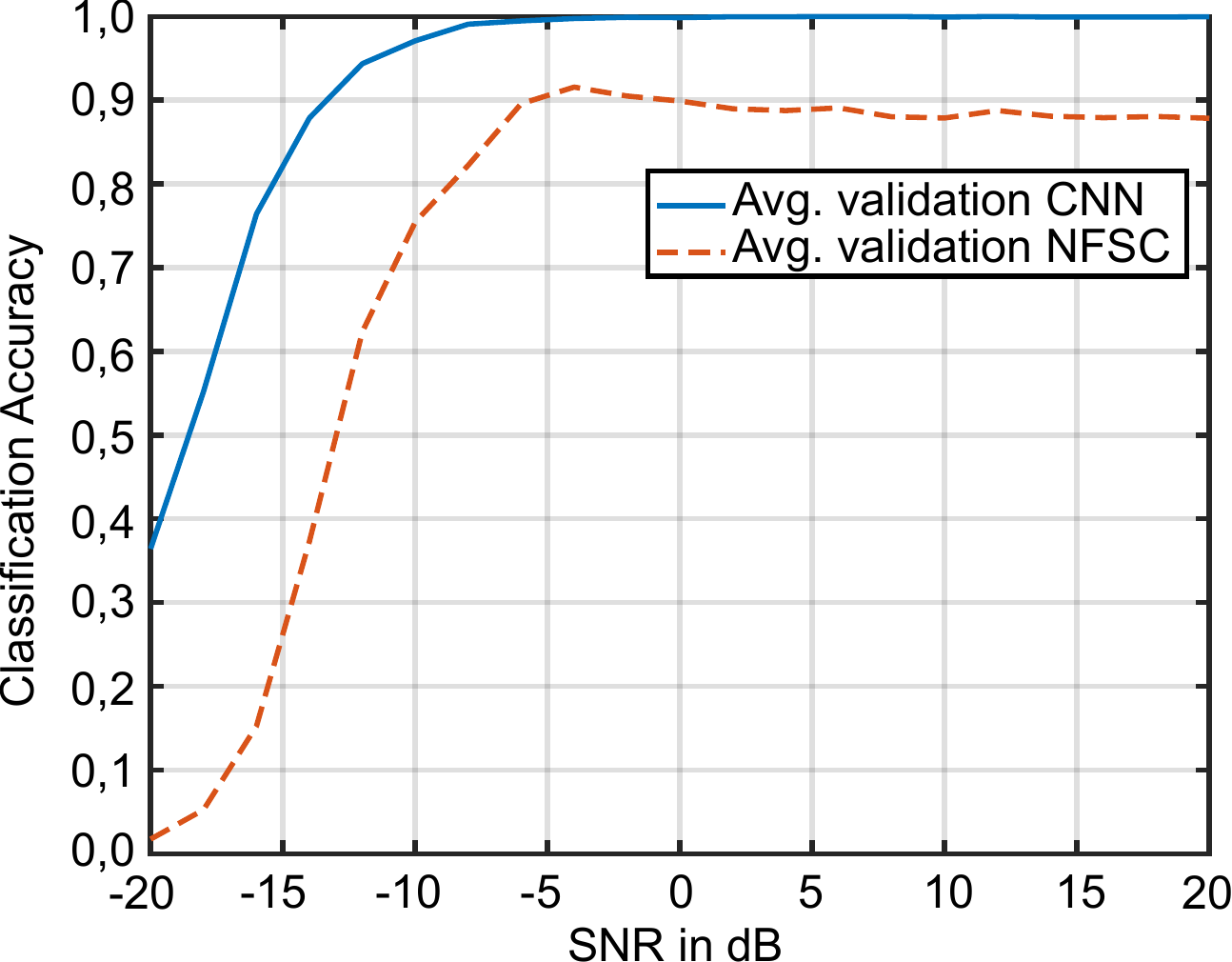}
	\caption{Comparative performance evaluation of \ac{CNN} and \ac{NFSC} with 
		averaged classification accuracy for IEEE 802.15.1 and IEEE 802.15.4 complaint signals}
	\label{fig:cnn_nfsc}
\end{figure}

The performance under such constraints of the \ac{NFSC} 
in comparison with \ac{CNN} is shown in Fig.~\ref{fig:cnn_nfsc}. 
It is important to note, that the resulting classification accuracies 
are averaged for all utilized classes.
For the applied validation data, the \ac{CNN} outperforms the \ac{NFSC} 
in terms of classification accuracy independent of \acp{SNR}. 
The \ac{CNN} shows an average performance gain and classification accuracy improvement 
of at least \unit[5.32]{dB} and \unit[8.19]{\%}, respectively.
Hence, the data-driven \ac{CNN} approach is limited to sensing snapshots similar to the trained data, 
while the \ac{NFSC} can be also utilized for real world scenarios.
Nevertheless, \ac{CNN}'s promising results should be investigated further. 

\subsection{Frequency- and Time-Domain Sensing Snapshots}
The \ac{CNN} can either be trained with sensing snapshots in time- or in 
frequency domain. 
Hence, the training was also performed with same-sized time-equivalent with 
quadrature and imaginary components of the raw \unit[128]{IQ-Samples} 
as  $128 \times 2$ input matrix.

\begin{figure}[htb]
	\centering
	\includegraphics[width=\columnwidth]{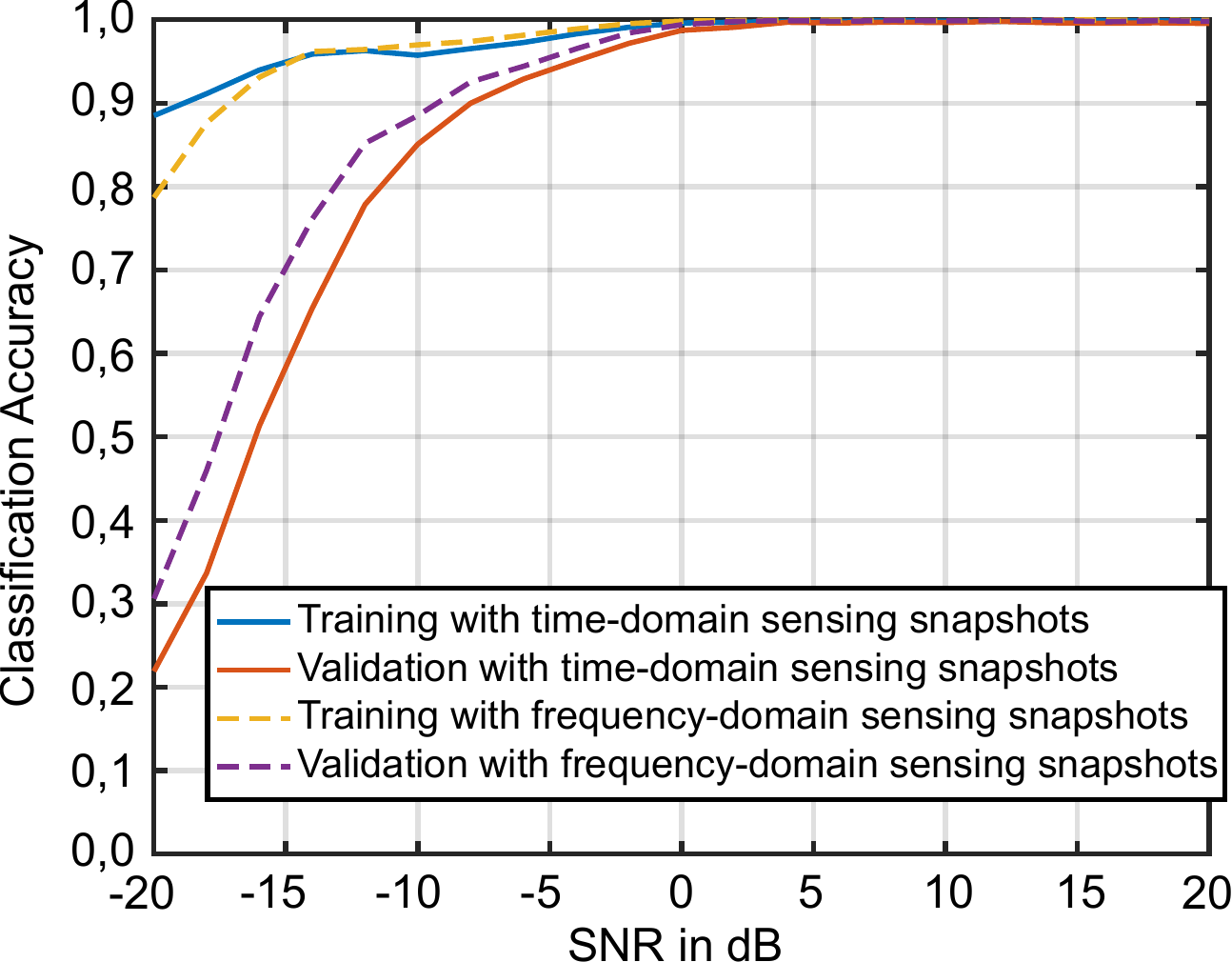}
	\caption{Comparative performance evaluation of \ac{CNN} trained with 
		sensing snapshots in time- and in frequency-domain}
	\label{fig:time_frequency}
\end{figure}

The classification accuracy with both frequency- and time-Domain sensing snapshots 
as input are shown in  Fig.~\ref{fig:time_frequency}
As expected the frequency-domain sensing snapshots outperform the time-equivalent
because foremost the \acp{CNN}'s differentiation has to be frequency selective.

\subsection{Reduced \ac{CNN}}
\label{sec:big_small_cnn}
\begin{figure}[htb]
	\centering
	\includegraphics[width=\columnwidth]{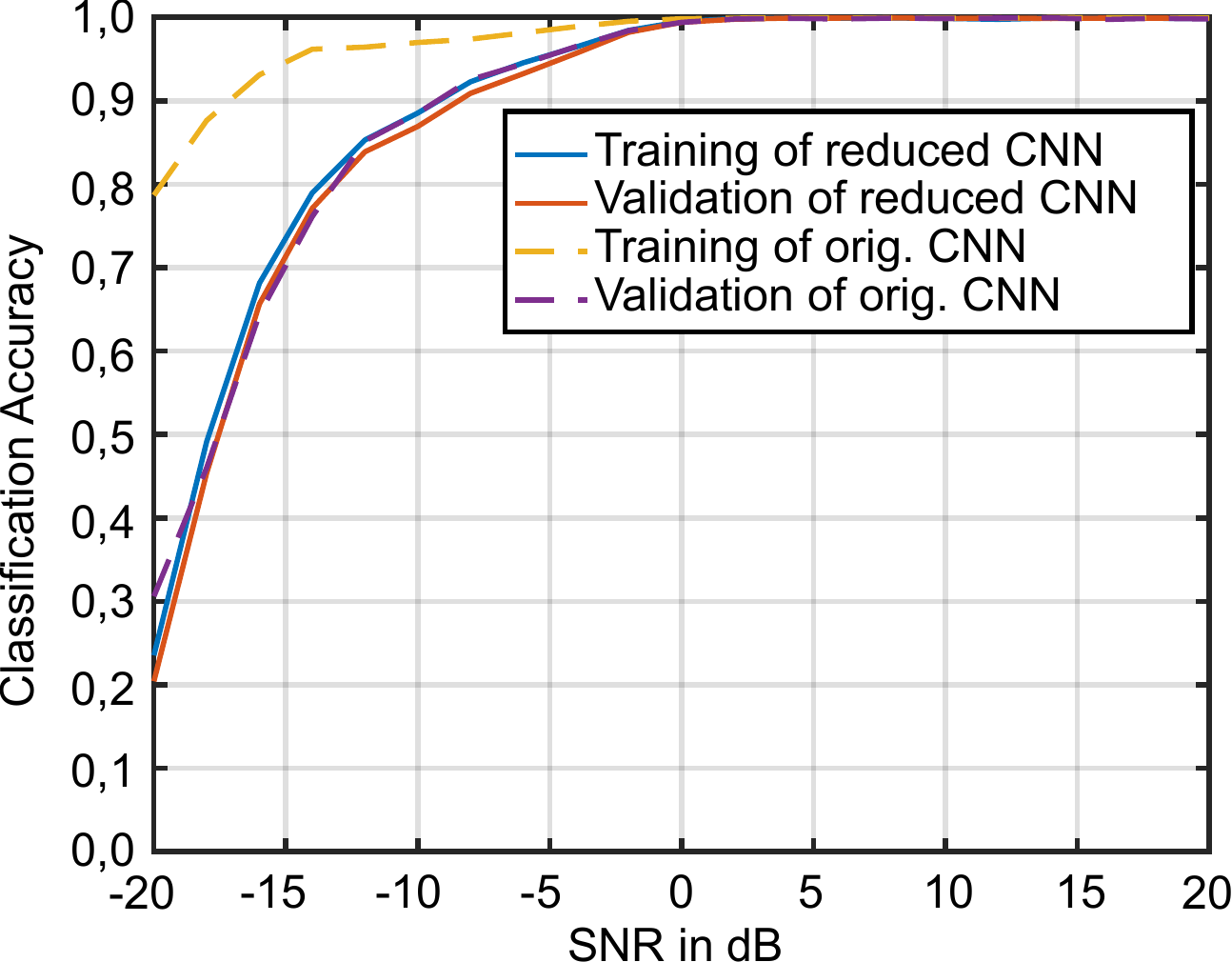}
	\caption{Classification accuracy of the original compared to the reduced \ac{CNN}}
	\label{fig:big_small_cnn}
\end{figure}

The reduced \ac{CNN} has over $99\%$
reduced degrees of freedom compared to the big \ac{CNN} and is therefore much
faster. The accuracy of the training and validation data of both \acp{CNN} is 
shown in Fig.~\ref{fig:big_small_cnn}.

Although the degrees of freedom for the small \ac{CNN} were reduced
significantly the accuracy of the validation data is similar. 
While for training the classification accuracy drops for the reduced \ac{CNN}
especially for low \acp{SNR} values,
for validation the performance of both \acp{CNN} is comparable.
The training performance difference shows that the original \ac{CNN} 
memorizes the sensing snapshot pattern including noise 
while the reduced \ac{CNN} generalizes much better.

%% file: conf_conclusion.tex
\section{Conclusion}
\label{sec:conclusion}

The steadily growing use of license-free frequency bands requires reliable 
coexistence management and therefore proper \ac{WII}.

In this work we propose the first \ac{WII} approach based upon deep \ac{CNN}.
The \ac{CNN} naively learns its features through self-optimization 
during an extensive data-driven training process.
The design of the \ac{CNN} was derived from the network 
of O'Shea et al. \cite{2016.OSheaCorgan} for the related field of modulation recognition.
Further, a network size reduction was performed by \unit[99]{\%} of the total degrees of freedom.
During the training phase it showed better generalization.

In order to face realistic wireless device capabilities the \ac{CNN} identifies
time- and frequency-limited sensing snapshots 
with a duration of \unit[12.8]{\textmu s} 
and acquisition bandwidth of \unit[10]{MHz}.
Thereby, it differs between 15 classes,
which represent allocated frequency channels 
of IEEE 802.15.4, IEEE 802.15.1, 
and partly in-band IEEE 802.11 b/g compliant packet transmissions.
In total 151,200 sensing snapshots were used for training and 74,025 for 
validation containing 19 different variants of modulation types and symbol rates.

For implementation the Python library Keras in combination with Theano 
was utilized with a high level of abstraction.
The GPU-based training speeds up the training duration 
to approximately \unit[501]{ms} for the original and \unit[20]{ms} for the reduced \ac{CNN} 
per batch size of 1024.

The proposed \ac{CNN} shows  promising results with high classification 
accuracy for signals with low \ac{SNR}. 
In average, the accuracy exceeds \unit[95]{\%} for \acp{SNR} greater than \unit[-5]{dB}.
The performance drops with wideband signals which are clipped 
by the limited acquisition bandwidth such as IEEE 802.11 b/g compliant packet 
transmissions.
Secondly, it also shows minimal performance issues 
with transmission signals sharing the same center frequency 
such as the fourth IEEE 802.15.1 and first IEEE 802.15.4 channel.
Nevertheless, the \ac{CNN} outperforms state-of-the-art \ac{WII} 
approaches such as \ac{NFSC} \cite{2010.Ahmad} under similar constraints.
Thereby, the \ac{CNN} approach shows an average performance gain 
and classification accuracy improvement 
of at least \unit[5.32]{dB} and \unit[8.19]{\%}, respectively.

Nevertheless, the prototype has to be enhanced
and must be validated in the field to become a viable option as a classifier
for coexistence management. 
Such an enhancement is a design of a \ac{CNN} suitable for multi-label \ac{WII} 
of concurrent transmissions in the same frequency range.
Secondly, the training data has to be extended and diversified to get a
better representation of channel and hardware impairments for the training data.